# Identifying 3 moss species by deep learning, using the "chopped picture" method

Takeshi Ise[1*], Mari Minagawa[2], and Masanori Onishi[3]

[1] Field Science Education and Research Center, Kyoto University, Japan
[2] Faculty of Agriculture, Kyoto University, Japan
[3] Graduate School of Agriculture, Kyoto University, Japan

* corresponding author: ise@kais.kyoto-u.ac.jp

**Abstract**

In general, object identification tends not to work well on ambiguous, amorphous objects such as vegetation.  In this study, we developed a simple but effective approach to identify ambiguous objects and applied the method to several moss species.  As a result, the model correctly classified test images with accuracy more than 90%.  Using this approach will help progress in computer vision studies.

**Introduction**

Especially in recent years, deep learning has become a very effective tool for object identification (Krizhevsky et al. 2012, Szegedy et al. 2015).  However, deep learning is mostly applied for distinctive and deterministic objects such as human face, human body, cat, dog, automobile, etc.  On the other hand, ambiguous objects such as trees, shrubs, and herbs are not very suitable for identification with machine learning.

In this study, we applied a new method called "chopped picture" to identify ambiguous, amorphous objects.  In our case study, mosses (bryophytes), a type of green plants, are the target objects for identification.  In general, differing from animals, plants have characteristics of modular growth.  For example, numbers of leaves and stems which one individual typically possesses are not predetermined but change flexibly according to environmental conditions.  This is clearly different that the standard model of human face has predetermined numbers of parts, such as two eyes, one mouth, and so on.  This characteristics of plants can make object identification difficult.

Moreover, differing from flowering plants, mosses do not have vascular organs and their tissues are not well differentiated, and thus mosses are even more difficult target for machine learning.  In addition, mosses tend to form mats consisted by numerous individuals of the same species.  This behavior is challenging for object identification because it is very difficult to distinguish individuals.  We willfully selected this difficult target for this case study, in order to test whether the "chopped picture" method can overcome the difficulties.

**Methods and Results**

Kyoto, Japan is famous for beautiful moss gardens because its wet and moderately warm climate is suitable for moss growth.  The study site is Murin-an, a traditional Japanese garden owned by Kyoto municipal government (Fig. 1).  In this study, we selected 3 moss genus (*Polytrichum* species (POL), *Trachycystis* species (TRA), and *Hypnum* species (HYP)) for identification.  In addition to these 3 categories, we have another visual category "not moss (NOM)."

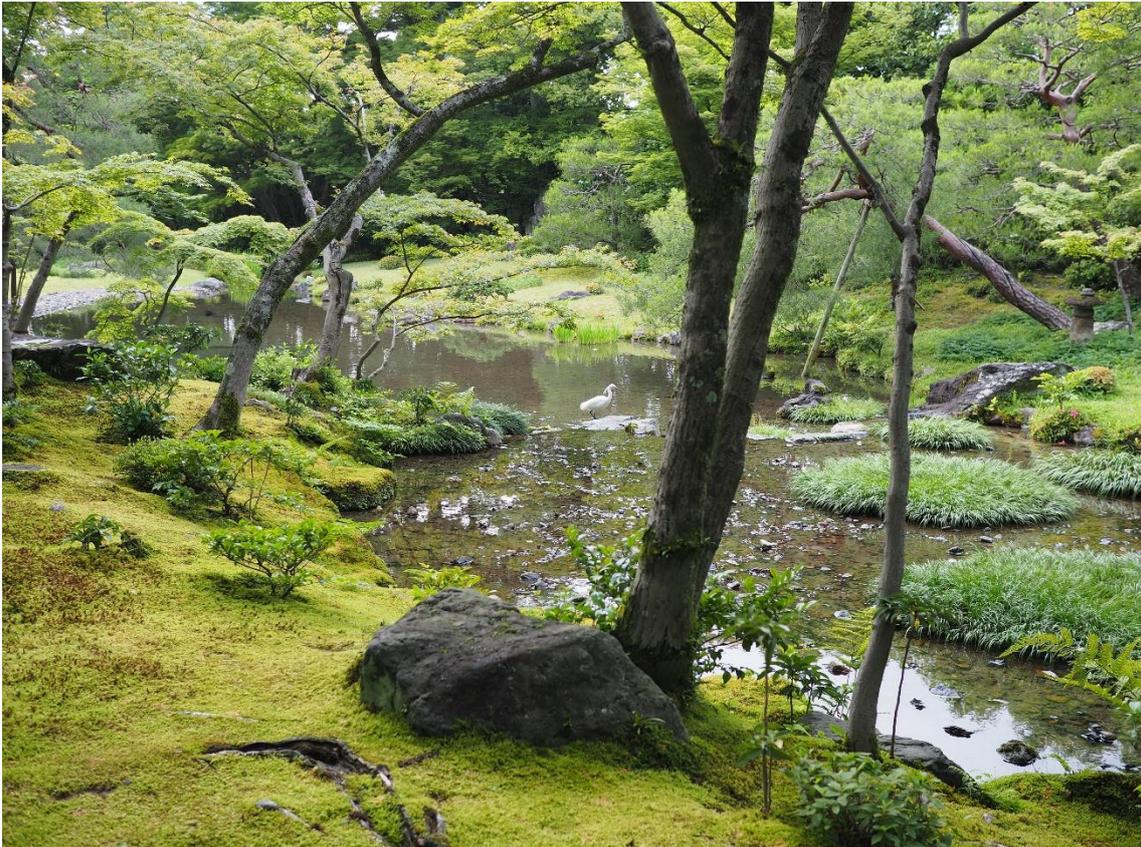

Fig. 1.  The scenery of Murin-an garden.  The ground of this Japanese garden, the moisture level is relatively high due to rainfall and nearby ponds, and sunlight is moderately available due to sparsely planted trees.  These environmental conditions make this garden a good habitat for several moss species.

The training data is prepared by the "chopped picture" method.  First, in this method, we found monotone moss patches, which are uniformly covered by a single moss species, and took their pictures with digital camera (Fig. 2).  For example, the picture showing in Fig. 2 has a pixel size of 4608 x 3456.  Next, using R 3.3.2 (R Core Team 2016), we "chopped" this picture into small squares (56 x 56) with 50% overlap both vertically and horizontally (Fig. 3).  With this method, it is very easy to obtain the training data because we simply need only a few digital photographs to make thousands of training data.  We used Olympus OM-D E-M5 Mark II (16 mega pixels) and Panasonic LUMIX G 20mm/F1.7 II ASPH (focal length of this lens is equivalent to 40 mm in 35 mm film), and took photographs directly above the moss patches by 60 cm.  We repeated this protocol for the 3 moss types (POL, TRA, and HYP) and "not moss" category (NOM).

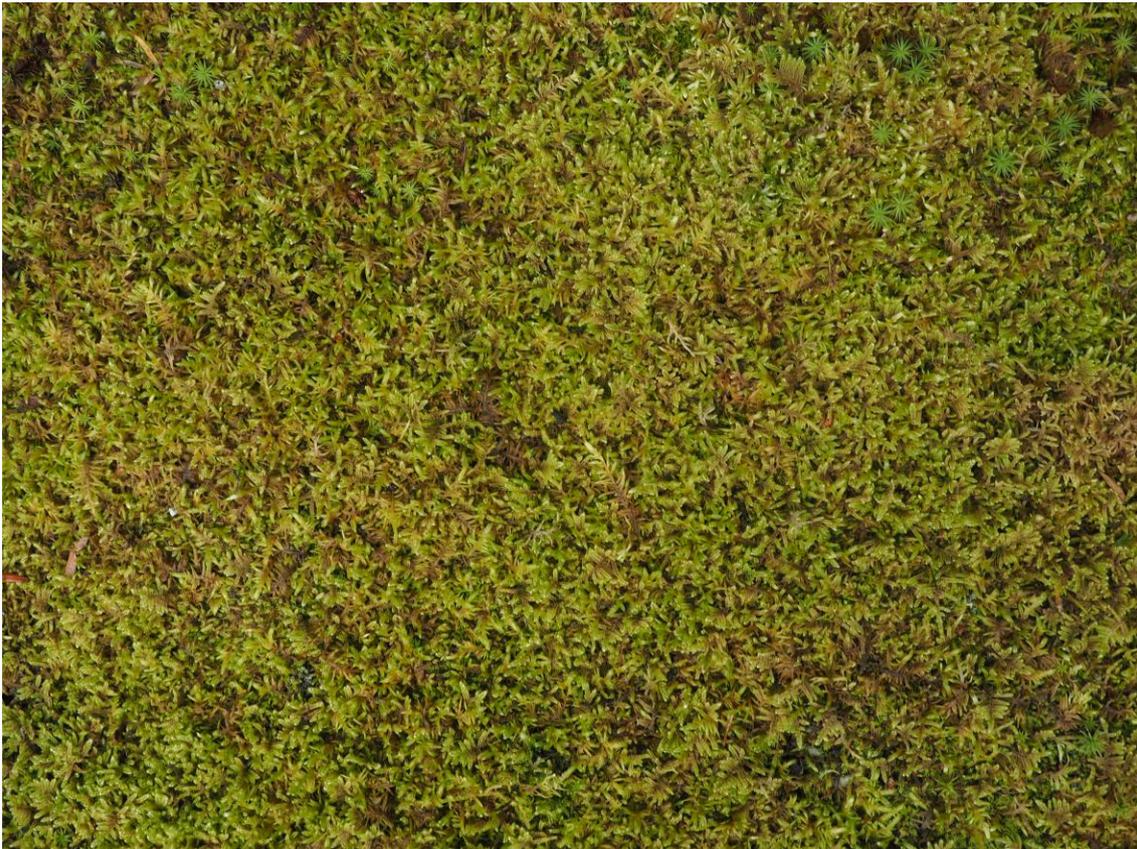

Fig. 2. A photograph of the ground of Murin-an garden.   Almost all area in this photograph is covered by *Hypnum* species (HYP).    By chopping this large digital image (4608 x 3456 pixels) into small squares (56 x 56 pixels) with 50-% overlap, we were able to obtain a large training dataset for deep learning easily.    This mat of *Hypnum* species has a small fraction of contamination with *Polytrichum* species.    The small squares with the impurities are screened and removed "by hand."

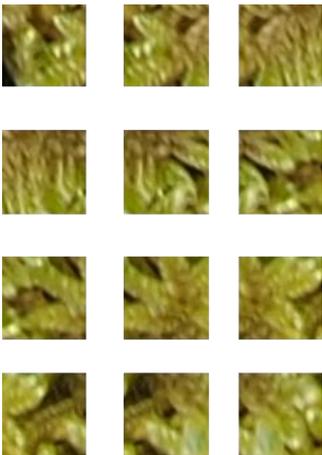

Fig. 3. A set of examples of small squares (56 x 56 pixels) of the photograph of *Hypnum* species (HYP).

To make a model for object identification, we used the deep learning framework of nVIDIA DIGITS 4.0 (Barker & Prasanna 2016) running on Ubuntu 14.04.    The computer used is a BTO machine with Xeon E5-2603v3 (1.60 GHz, 6 cores), 16 GB RAM, and nVIDIA Quadro

K620. The summary of the training is shown in Fig. 4. In total, we obtained 93,851 images for training and used 25 % for validation. We used the LeNet network model. Other learning settings are basically the default to the DIGITS Image Classification Model (Table 1).

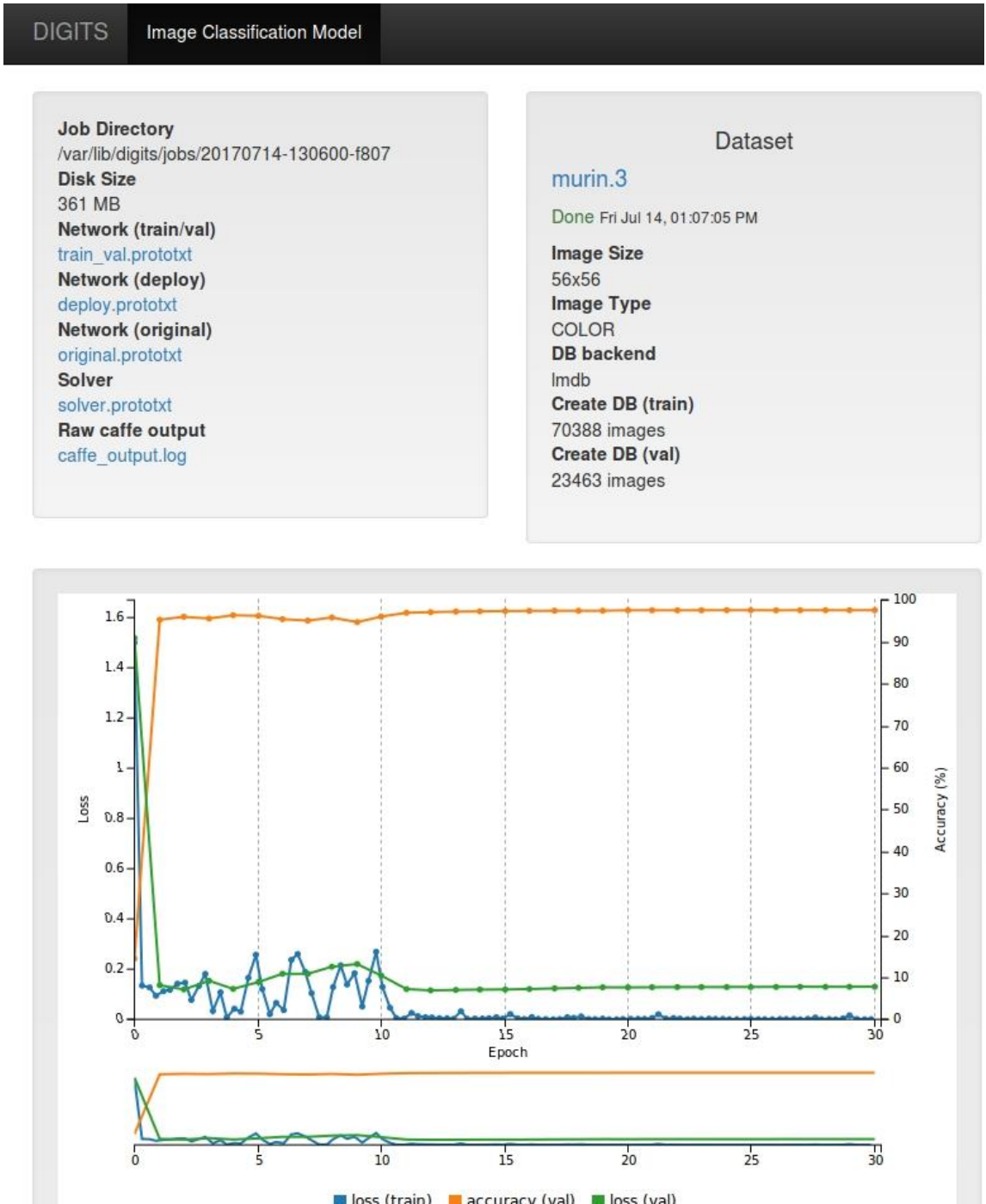

Fig. 4. Summary of training for object identification of 4 categories of moss garden.

Table 1. The setting of the DIGITS Image Classification Model.

| Setting | Selected option |
|---|---|
| Training epochs | 30 |
| Snapshot interval | 1 |
| Validation interval | 1 |
| Random seed | None |
| Batch size | Network defaults |
| Solver type | SGD |
| Base learning rate | 0.01 |
| Policy | Step down |
| Step size | 33 |
| Gamma | 0.1 |
| Network | LeNet |

To test the trained model, we selected a picture which is also taken with the same protocol (focal length of 50 mm and 60 cm above the surface). In this picture, the target moss species and non-moss objects are included (Fig 5). To test the model, we also "chopped" this picture in the same manner as the training data. Using the model testing function of DIGITS 4.0, the model gives prediction and confidence of that prediction. Then we re-organize the small pieces into the original picture and showing predicted categories with colored circles (Fig 5). In this result, the model classified moss species appropriately. In the bottom right, the picture is mostly covered uniformly by POL, and the model nicely classified the objects. There are some growth of HYP within the POL patch, and the model has appropriately found HYP individuals.

The classification of TRA (mainly in the middle) and HYP (mainly in the left) are generally appropriate, but with some classification errors. To obtain quantitative performance, we arbitrarily set regions where one moss species dominate. In those regions, from top left corner, we locate circles with the color of the dominated species, then evaluate whether the model prediction is correct or incorrect. We repeated this for 100 times for each region, and calculated the performance. The estimated performance for POL is 99%, TRA is 95%, and HYP is 74%.

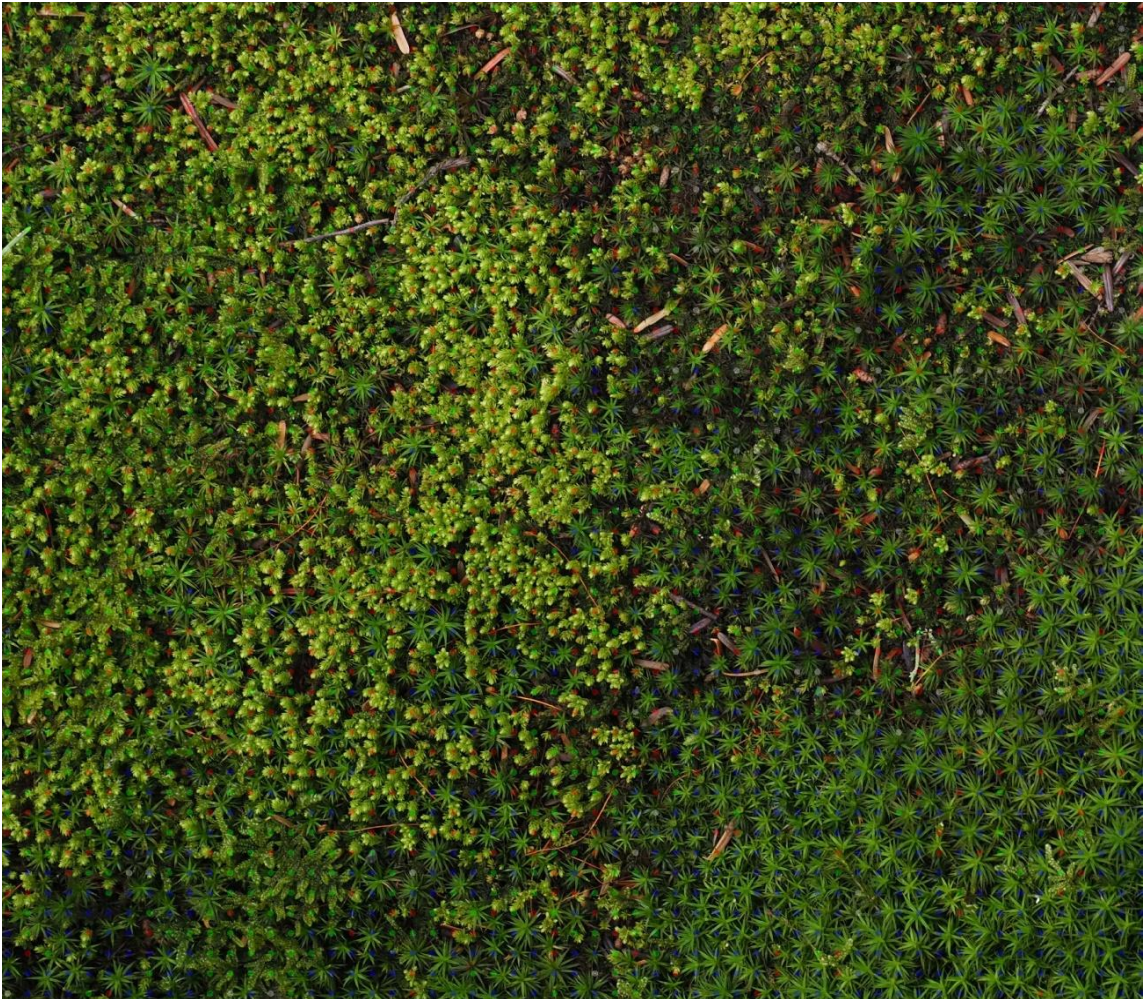

Fig. 5. An example of model performance test with 3 moss species and "not moss" categories.  The results of classification are color-coded (blue: POL, red: TRA, green: HYP, and white: NOM).   The original image can be obtained at <https://www.dropbox.com/s/hm6a5kjofnm1dsv/Clipboard01.jpg>.

**Discussion**

Our "chopped picture" method can be a good way to identify ambiguous objects such as green plants, especially mosses.   Although mosses are highly amorphous, our method generally performed well.

However, the model performance was not the same for the 3 moss species.   Identification of POL is excellent because this moss species are relatively large and has relatively distinctive, well-defined shape.   On the other hand, the performance for HYP was not very good because this moss species is highly amorphous; it strongly shows vegetative growth with runners.   As the result, the shape of HYP patches become amorphous and this characteristics affected the model performance.

To improve the model, we have the following suggestions.   Color standardization can improve the performance.   Color standardization using a color chart for photography will allow us to standardize the white balance and exposure.   A trial and error for the "chopping size" should also be performed for the better performance.   Although mosses are evergreen plants, they still show some seasonal differences in color, shape, and size.   For example, spring is a growing season and TRA produces light green parts, easily visible for human eyes, in this season.   Moreover, mosses produce reproductive organs in particular timings.   Thus, collecting training data in many seasons and situations can improve the model.

Our results suggest several implications.   Using the "chopped picture" method, we show that the performance of object identification with deep learning can be applied to things with amorphous shapes.   This method can be applied for visual observations of plants in various scales, such as field photography (in this study), drone photography, aircraft photography such as Google Earth, and satellite images.   In addition, as a practical application, we hope to make an application of "moss identifier" that allows people to know the moss names using a smartphone.

## Acknowledgements

This work was a part of the Utilizing the Knowledge of Universities: a Multilateral Municipal Government Research Project, Kyoto City, Japan, and also supported by the JST PRESTO Grant Number JPMJPR15O1, Japan.   We deeply appreciate professional advice from Drs. Yoshitaka Oishi and Wakana Azuma.